\documentclass[10pt, conference, compsocconf]{IEEEtran}
\usepackage{float} 
\usepackage{cite}
\ifCLASSINFOpdf
  \usepackage[pdftex]{graphicx}
  \DeclareGraphicsExtensions{.pdf,.jpeg,.png}
\else
  \usepackage[dvips]{graphicx}
  \DeclareGraphicsExtensions{.eps}
\fi
\usepackage[cmex10]{amsmath}
\usepackage{algorithmic}
\usepackage{array}

\ifCLASSOPTIONcompsoc
 \usepackage[caption=false,font=normalsize,labelfont=sf,textfont=sf]{subfig}
\else
 \usepackage[caption=false,font=footnotesize]{subfig}
\fi
\usepackage{fixltx2e}
\usepackage{stfloats}
\usepackage{url}
\usepackage{amssymb}
\usepackage{gensymb} 
\usepackage{multirow} 

\usepackage{color}

\begin{document}
\title{A Multi-task Deep Learning Architecture for Maritime Surveillance using AIS Data Streams}

\author{\IEEEauthorblockN{Duong Nguyen\IEEEauthorrefmark{1},
Rodolphe Vadaine\IEEEauthorrefmark{2},
Guillaume Hajduch\IEEEauthorrefmark{2}, 
Ren\'e Garello\IEEEauthorrefmark{1} and
Ronan Fablet\IEEEauthorrefmark{1}}
\IEEEauthorblockA{\IEEEauthorrefmark{1}IMT Atlantique, Lab-STICC, UBL, 29238 Brest, France\\ Email: \{van.nguyen1, rene.garello, ronan.fablet\}@imt-atlantique.fr}
\IEEEauthorblockA{\IEEEauthorrefmark{2}CLS - Space and Ground Segments, 29280 Brest, France \\
Email: \{rvadaine, ghajduch\}@cls.fr}
}

\maketitle

\begin{abstract}

In a world of global trading, maritime safety, security and efficiency are crucial issues. We propose a multi-task deep learning framework for vessel monitoring using Automatic Identification System (AIS) data streams. We combine recurrent neural networks with latent variable modeling and an embedding of AIS messages to a new representation space to jointly address key issues to be dealt with when considering AIS data streams: massive amount of streaming data, noisy data and irregular time-sampling. We demonstrate the relevance of the proposed deep learning framework on real AIS datasets for a three-task setting, namely trajectory reconstruction, anomaly detection and vessel type identification.


\end{abstract}
\IEEEpeerreviewmaketitle

\section{Context}
\label{secContext}

In the world of a globalized economy, maritime surveillance is a vital demand. Currently being the most efficient long-distance transporting method, sea transport is carrying about 90\% of the world trade\footnote{According to the International Maritime Organization (IMO)}. With the persistent growth of maritime traffic, safety and security are key issues. Besides, the real-time delivery of maritime situation maps is also necessary for a variety of activities: fishing activities control, smuggling detection, EEZ intrusion detection, transshipment detection, maritime pollution monitoring, etc.

Over the last decades, the development of terrestrial networks and satellite constellations of Automatic Identification System (AIS) has opened a new era in maritime traffic surveillance. Every day, AIS provides on a global scale hundreds of millions of messages \cite{perobelli_marinetraffic-day_2016}, which contain ships' identifier, their Global Positioning System (GPS) coordinates, their speed, course, etc. The potential of this massive amount of data is clearly of interest if tools and models provide means to efficiently extract, detect and analyze relevant information from these data streams. However, current operational systems, which strongly rely on human experts, can only deal with a limited fraction of AIS data. 
 
Thus, the development of AI-based systems is a critical challenge. Beyond the volume of streaming data to be dealt with, there are two other key issues make it difficult to design these types of systems: noise patterns exhibited by AIS data and the irregular time-sampling. 
Both are very common in AIS and make the direct application of state-of-the-art supervised machine learning models, including deep learning ones poorly adapted.

This paper addresses these issues and explores deep learning models and architectures, and more specifically Recurrent Neural Networks (RNNs) to develop an automatic system that can process and detect, extract and characterize useful information in AIS data streams for maritime surveillance. More specifically, our key contributions are three-fold: 
\begin{itemize}
\item The design of a novel big-data-compliant unsupervised architecture which automatically learns and extracts useful information from noisy and partial AIS data streams on a regional scale;
\item The joint exploitation of this architecture as a basis for specific tasks using mathematically-sound statistical models, namely trajectory reconstruction and forecasting, maritime route estimation, vessel type identification, detection of abnormal vessel behaviors, etc.; 
\item The demonstration of the proposed approach's  relevance on real regional datasets off Brittany coast and in the Gulf of Mexico, significantly more complex than case-studies addressed in previous works. 
\end{itemize}

This paper is organized as follows: in Section \ref{secRelatedWork}, we review the state-of-the-art methods in AIS-based maritime surveillance. Section \ref{secBackground} provides the background to understand this paper. The proposed method is detailed in Section \ref{secModelsArchitecture}. We present experiments in Section \ref{secExperimentsResults}, and further discuss the main features and performance of our approach in  Section \ref{secExplanations}. Finally, conclusions and perspectives for future work are presented in Section \ref{secConclusionsPerspectives}.  

\section{Related work}
\label{secRelatedWork}

In this section, we review the related works in the field of AIS-based maritime traffic surveillance, especially regarding trajectory reconstruction and forecasting and anomaly detection.

\textbf{Trajectory reconstruction and forecasting}: 
For simplicity purpose, we use here the term ``trajectory reconstruction'' to refer to both trajectory reconstruction and trajectory forecasting. Early efforts for trajectory reconstruction include linear interpolation, curvilinear interpolation \cite{best_new_1997} and their improvements \cite{perera_maritime_2012}, \cite{schubert_comparison_2008}. They rely on a physical model of the movement $x_t = x_{t-1} + \Delta t*x'_t$ (where $x_t$ is the position of vessel at the time $t$, $x'_t$ is the deviation of $x_t$, usually the Speed Over Ground (SOG) and the Course Over Ground (COG)). More sophisticated methods suppose that vessel trajectories follow a distribution and learn it from historical data \cite{millefiori_modeling_2016}, \cite{pallotta_context-enhanced_2014}. Currently, state-of-the-art methods for trajectory reconstruction \cite{mazzarella_knowledge-based_2015}, \cite{hexeberg_ais-based_2017}, \cite{coscia_multiple_2018} 
use the following typical three-step approach: i) the first step involves a clustering method, {\em e.g.} TRACLUS \cite{lee_trajectory_2007} or TREAD \cite{pallotta_vessel_2013} to cluster historical motion data into route patterns, ii) the second one assigns the vessel to be processed to one of these clusters iii) the third one interpolates or predicts the vessel trajectory based on the route pattern of the assigned cluster. 

\textbf{Anomaly detection}: Some models detect abnormal behaviors by defining them explicitly \cite{holst_stattistical_2016}, \cite{gaspar_analysis_2016}. These types of models  are usually limited themselves by their own definitions, and can not handle all the complex phenomenons observed at sea. To overcome those drawbacks, other methods detect anomalies implicitly by creating normalcy models, then consider trajectories or trajectory segments that do not suit these models as abnormal. In \cite{rhodes_maritime_2005}, Rhodes divided the map into small zones and used Normalcy Box to detect abnormal vessel speed in each zone. 
Gaussian Mixture Models (GMMs), Kernel Density Estimation (KDE) were explored in \cite{laxhammar_anomaly_2008}, \cite{ristic_statistical_2008}. More sophisticated methods have used time series analysis techniques, such as Gaussian Process \cite{kowalska_maritime_2012}, \cite{will_fast_2011}, or Bayesian Networks (BNs) \cite{johansson_detection_2007}, \cite{mascaro_anomaly_2014} to capture the sequential structure of AIS streams. All these models share the same basic idea: in a small region, vessels should perform similar behaviors. 

All models and approaches reviewed for trajectory reconstruction and anomaly detection present three main drawbacks:
\begin{itemize}
	\item They depend on strong priors and can hardly capture all the heterogeneous characteristics of AIS data as well as the varieties of vessels' behaviors. 
Near-far, fast-slow, etc. are relative definitions and are difficult to be implemented. Almost all current models work only for cargo and tanker vessels on specific high-traffic maritime routes. However, more sophisticated models and relaxed assumptions are required to address the range of vessel types and vessel behaviors revealed by AIS streams on a regional or global scale. 
	\item Most if not all methods exploit at some point a clustering. They typically assume that in specific areas, all vessels tend to perform similar behaviors, and then use clustering methods (Kmeans, DBSCAN, etc.) to find those behaviors. For example, for trajectory reconstruction issues, each cluster is a maritime route \cite{pallotta_vessel_2013}; in anomaly detection, each cluster is a speed mode \cite{pallotta_context-enhanced_2014}, \cite{rhodes_maritime_2005}, etc. We believe that such clustering steps result in information losses. By contrast, we argue that continuous latent states should be preferred 
to address the complexity of AIS data streams.
	\item Current methods do not explicitly address the  irregular time-sampling of AIS streams. Non-sequential methods \cite{rhodes_maritime_2005} do not take it into account and sequential ones \cite{mazzarella_knowledge-based_2015} assume they are provided with regularly-sampled streams, which is not true or may result in the creation of artificial, possibly erroneous AIS positions if interpolation techniques are used as a pre-processing step.
\end{itemize}

As detailed hereafter, we develop a novel multi-task deep learning framework to address these issues and demonstrate its relevance from experiments on a real AIS dataset on a regional scale.


\section{Recurrent neural networks with latent variables}
\label{secBackground}
Deep learning has rapidly become the state-of-the-art framework for a wide range of machine learning problems \cite{lecun_deep_2015}, especially when dealing with large-scale datasets. In this context, recurrent neural networks have showed their advanced abilities for time series processing. Recently, the introduction of recurrent neural networks with latent variables \cite{chung_recurrent_2015}, \cite{fraccaro_sequential_2016}, \cite{maddison_filtering_2017} has further extended the application range of RNNs to deal with noisy and heterogeneous sequential data modeling. Latent variables in RNNs provide this type of networks the ability to infer rich state representations and flexible non-linear transition models to account for complex randomness sources \cite{chung_recurrent_2015}.


The aim of recurrent neural networks with latent variables is to model the distribution $p$ of  a sequence of $T$ observed random variables $x_{t, t = 1..T}$. We assume that the generation process of $\{\mathbf{\mathit{x_t}}\}$ relies on a sequence of $T$ latent variables $z_{t, t = 1..T}$. The joint distribution $p(x_{1:T},z_{1:T})$ could be factored into:
\begin{equation}
p(x_{1:T},z_{1:T}) = p_1 (x_{1},z_{1})\prod_{t=2}^T p_t(x_t,z_t|x_{1:t-1},z_{1:t-1})
\label{eqJoinDistribution}
\end{equation}
$p_t(x_t,z_t|x_{1:t-1},z_{1:t-1})$ is then factored into two conventional steps of hidden state space models where the prior distribution $p(z_t|x_{1:t-1},z_{1:t-1})$ denotes the transition function and the  conditional distribution $p(x_t|x_{1:t-1},z_{1:t})$ denotes the emission function. The distribution of the observed data $p(x_{1:T})$ is given by: 
\begin{equation}
  p(x_{1:T}) = \mathbb{E}_{z_{1:T}} \big[ p(x_{1:T},z_{1:T}) \big] 
  \label{eqDataDistribution}
\end{equation}

However, the integral over $z_{1:T}$ in Eq. \ref{eqDataDistribution} can not be calculated directly. To circumvent this obstacle, the most common approach \cite{chung_recurrent_2015}, \cite{fraccaro_sequential_2016}, \cite{maddison_filtering_2017} is to introduce an approximation $q(z_t|x_{1:t},z_{1:t-1})$ of the posterior distribution $p(z_t|x_{1:t},z_{1:t-1})$ then estimate $p(x_{1:T})$ by the Evidence Lower BOund (ELBO):
\begin{multline}
	\log p(x_{1:T}) \geq \mathcal{L}(x,p,q) = \mathbb{E}_{z_{1:T}\sim q}\big[\log p(x_{1:T}|z_{1:T}) \big] \\
    - \mathcal{KL}\big[q(z_{1:T}|x_{1:T})||p(z_{1:T}) \big] 
	\label{eqELBO}
\end{multline}
where $\mathcal{KL}\big[q||p\big] $ is the Kullback-Leibler divergence between two distributions $q$ and $p$. 

This estimation can be applied for all the family of $p$ and $q$ that can be factored over t:

\begin{equation}
	p(x_{1:T}|z_{1:T}) = \prod_{t=1}^T p(x_t|x_{1:t-1},z_{1:t})
\end{equation}

\begin{equation}
	q(z_{1:T}|x_{1:T}) = \prod_{t=1}^T q(z_t|x_{1:t},z_{1:t-1})
\end{equation}

\begin{equation}
	p(z_{1:T}) = \prod_{t=1}^T p(z_t|x_{1:t-1},z_{1:t-1})
\end{equation}

In this paper, we use Variational Recurrent Neural Networks (VRNNs) of Chung et al. \cite{chung_recurrent_2015}. VRNNs model the historical information $(x_{1:t-1},z_{1:t-1})$ by the hidden state of its RNN $h_t = h_t(x_{t-1},z_{t-1},h_{t-1})$. The conditional distribution  $p(x_t|x_{1:t-1},z_{1:t}) = p(x_t|z_t, h_t)$, the prior distribution  $p(z_t|x_{1:t-1},z_{1:t-1}) = p(z_t|h_t)$ and the variational posterior distribution $q(z_t|x_{1:t},z_{1:t-1}) = p(z_t|x_t,h_t)$ are parameterized by fully connected networks.

\section{Proposed multi-task VRNN model for AIS data}
\label{secModelsArchitecture}

\begin{figure}[!t]
  \centering
  \includegraphics[width=85mm]{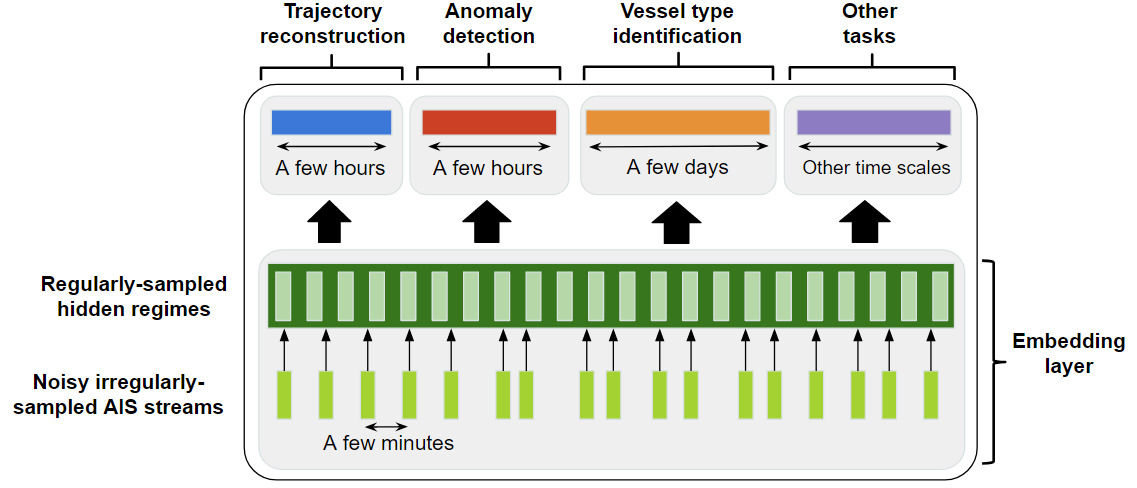}%
  \caption{Proposed VRNN architecture.} \label{figModelArchitecture}
\end{figure}

As sketched in Fig.~\ref{figModelArchitecture}, we propose a general multi-task Variational Recurrent Neural Network for the analysis of AIS data streams. Received AIS messages are regarded as irregular noisy observations of the true hidden states - called regimes; these regimes themselves may correspond to specific activities ({\em e.g.} under way using engine, at anchor, fishing, etc.). The key component of our model is the Embedding block, which converts noisy and irregularly-sampled AIS data to consistent and regularly-sampled hidden regimes.  This Embedding block relies on a VRNN \cite{chung_recurrent_2015} and operates at a 10-minute time scale. Higher-level blocks are  task-specific submodels, addressing at different time-scales ({\em e.g.} daily, monthly,...) the detection of abnormal behaviors, the automatic identification of vessel types, vessel position prediction, the identification of maritime routes, etc. 

\subsection{A latent variable model for vessel behaviors}

Through a VRNN architecture, we introduce hidden regimes (latents variables) as a data representation\footnote{Here we use the criteria defined in \cite{bengio_representation_2013} to evaluate this representation, readers are encouraged to read \cite{bengio_representation_2013} before continue.} that captures the true maneuvers of vessels (\textit{natural clustering}). Hidden regimes can be regarded as the ``roots'' of AIS messages. They govern how the vessel moves. 
From the point of view of higher levels (task-specific layers), hidden regimes provide the necessary information for their task (\textit{hierarchical organization of explanatory factors} and \textit{shared factors across tasks}). They disentangle the underlying information of AIS data (\textit{simplicity of factors dependencies}). For example, saying ``this vessel is performing a fishing maneuver'' is much more informative than saying ``the speed of this vessel is high''. 

It is important to note that the hidden regimes are  not clusters of AIS messages, because the act of assigning data to group would cause information loss. We share the same idea with \cite{kingma_auto-encoding_2013}, that latent variables (hidden regimes in this case) are continuous and there are no simple interpretations of these dimensions. 


The introduction of hidden regimes brings us two key benefits: an efficient encoding of AIS datasets and a regularly-sampled sequential representation. Regarding the first aspect, state-of-the-art systems such as TREAD \cite{pallotta_vessel_2013} have to store all the AIS messages in the training set, which is updated incrementally
new AIS messages. Therefore, data volume to be handled for the test phase increases rather linearly with the area of the Region Of Interest (ROI) and the duration of the considered time period. This may prevent such systems from scaling up to regional or global scales. By contrast, once the VRNN is trained, all the knowledge gained from a given AIS dataset is encoded by the characteristics of the hidden regimes, more precisely the fitted conditional distributions $p(z_t|x_{1:t-1},z_{1:t-1})$ and $p(x_t|x_{1:t-1},z_{1:t})$. 
Therefore, for the application of a trained model, there is no need to access the training dataset. This dataset may only be of interest to retrain or fine-tune a given model. It may be noted that the complexity of the representation of the hidden regimes ({\em i.e.}, the associated number of parameters) does not depend on the training data volume. For instance, in the considered experiments, for a dataset of more than $2.10^8$ AIS messages (each message contains several attributes), the fitted hidden regime representation involves about $5.10^6$  parameters. The second important feature is the mapping of an input space consisting of a noisy irregularly-sampled time series to a novel regularly-sampled sequential representation which naturally accounts for the different sources of uncertainties exhibited by AIS datasets. Hence, the proposed architecture embeds somehow a time regularization of the input data and does not require the definition of ad hoc denoising and interpolation pre-processing steps, which prove difficult due to the variabilities to be dealt with ({\em e.g.}, duration of the missing data segments, noise patterns, inhomogeneous space-time variabilities, etc.).
From a mathematical point of view, the considered model naturally embeds these issues through the time propagation of the approximate posterior distribution $q(z_t|x_{1:t},z_{1:t-1})$. Overall, this regularly-sampled sequential representation makes feasible the design of classic architectures on top of the embedding layer to deal with task-specific issues as detailed in Section \ref{secTrajectoryConstructionLayer}, \ref{secAbnomalyDetection} and \ref{secVesselTypeIdentification}.

\subsection{``Four-hot'' representation of AIS messages}
\begin{figure}[!t]
  \centering
  \includegraphics[width=85mm]{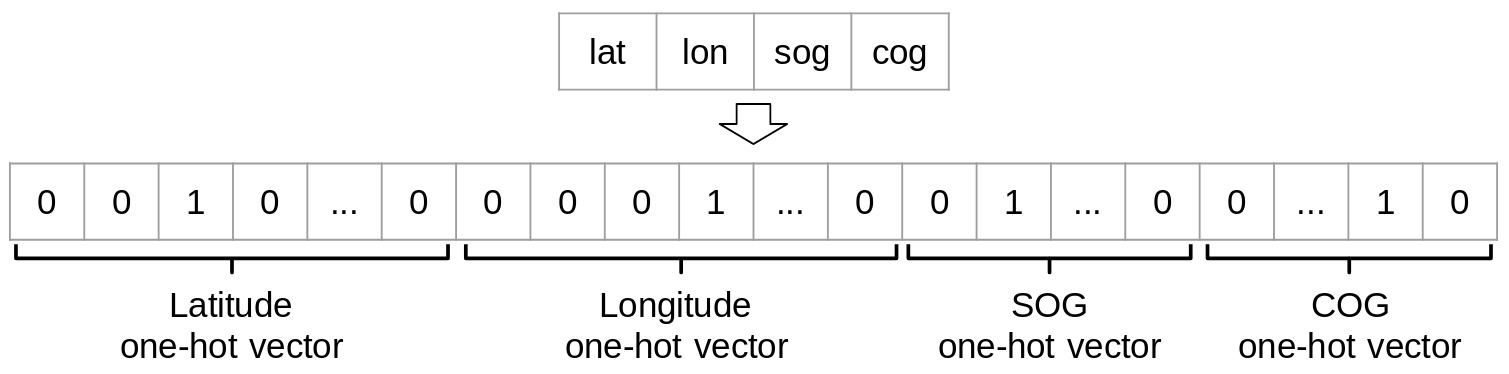}%
  \caption{``Four-hot'' vector.} \label{figFourHotVector}
\end{figure}

Instead of presenting AIS messages directly by their 4-D real-value vector $[lat, lon, SOG, COG]^T$ like methods in the literature \cite{pallotta_context-enhanced_2014}, we apply the bucketization technique to introduce a novel representation of AIS data: the ``four-hot encoding'' (Fig. \ref{figFourHotVector}). This representation, inspired by the one-hot encoding in language modeling, is created by concatenating the one-hot vectors of 4 attributes in AIS message: latitude coordinate, longitude coordinate, SOG and COG. To create the one-hot vector of an attribute, we simply divide the entire value range of this attribute into $N_{attribute\_i}$ equal-width bins. 

The ``four-hot encoding'' not only brings us the benefits of bucketized presentation but also provides a more structured representation to learn trajectory spatial patterns as illustrated in Section \ref{secExplanations}. Our four-hot representation shares similarities with \cite{jiang_trajectorynet:_2017}. However, in \cite{jiang_trajectorynet:_2017}, the authors explained their representation as a transformation from feature space to semantic space based on the smoothness prior assumption. They argued that the continuous values of features did not matter, the explanatory factors were the semantic interpretation presented in their discrete vector of these values. We, on the other hand, consider the ``four-hot encoding'' as a presentation that can i) accelerate the calculation of neural networks (similar to one-hot encoding), ii) disentangle some explanatory factors of input features (see Section \ref{secExplanations}). The semantic space in our architecture is the space of hidden regimes.    

The implicit reduction of the precision of the AIS position and velocity features may be regarded as a drawback of the four-hot representation. We however argue that for the targeted applications there is no need for the embedding block to provide precise numerical features. For example, a speed of 12 knots and a speed of 12.1 knots do not mean any difference in our context.

\subsection{Embedding block}
\label{secEmbeddingLayer}

The embedding block is a VRNN \cite{chung_recurrent_2015}, where $x_t$ is the ``four-hot encoding'' of AIS message and $z_t$ is the concatenation of the hidden state of the network and the latent variable at the time $t$. This layer works at a 10-minute time scale ({\em i.e.} we downsample AIS data stream to a resolution of 10 minutes) and learns the distribution $p(x_{1:T})$ (via the prior distribution $p(z_t|x_{1:t-1},z_{1:t-1})$, the generative distribution $p(x_t|x_{1:t-1},z_{1:t})$ and the approximative posterior distribution  $q(z_t|x_{1:t},z_{1:t-1})$).   

After being trained, the embedding layer consistently generates regularly time-sampled hidden regime series. This series is used as input to task-specific submodels as sketched in Fig.\ref{figModelArchitecture}. 

\subsection{Trajectory reconstruction submodel}
\label{secTrajectoryConstructionLayer}

The Embedding block is naturally a generative model, so the construction of a vessel trajectory estimator/predictor on top of this block is relatively direct. We follow the philosophy of \cite{mazzarella_knowledge-based_2015}. In this approach, one  infers maritime contextual information, which is used to enhance the prediction/estimation. The contextual information in \cite{mazzarella_knowledge-based_2015} was inferred by TREAD \cite{pallotta_vessel_2013}, which means that each vessel would be assigned to a predefined route. By contrast, we avoid such a hard assignment to a predefined behavioral cluster. We benefit from the richer contextual representation inferred by the Embedding block. Formally, the proposed trajectory reconstruction model is stated as the inference of the posterior $q(z_t|x_{1:t},z_{1:t-1})$ and the sampling-resampling from the distribution $p(x_{t+1}|x_{1:t},z_{1:t}) = \int p(x_{t+1}|x_{1:t},z_{1:t+1})p(z_{t+1}|x_{1:t},z_{1:t})dz_{t+1}$ (all learned by the Embedding block) using a particle filter \cite{maddison_filtering_2017}. 



\subsection{Abnormal behaviour detection submodel}
\label{secAbnomalyDetection}

The second specific task on top of the Embedding block is the detection of abnormal behaviors. It comes to define a normalcy model to detect the (unlikely) anomalies w.r.t. this model. As a direct by-product of the trained Embedding block, we can evaluate the likelihood $p(x_{1:T})$ of any input AIS sequence $x_{1:T}$ using a marginalization w.r.t. the hidden regimes. A series of AIS messages with a very low likehood w.r.t. a given threshold may be regarded as being unlikely for model $p(x_{1:T})$ and hence as abnormal. 

One may however consider context-aware detection rules. For example on maritime routes, vessels' behaviors are roughly identical, which leads to high values for the likelihood $p(x_{1:T})$. In other regions, the variety of vessel types and activities results in much more complex mixtures of behaviors and much lower likelihood values for the normalcy model. The selection of a global threshold over an entire region may not be as appropriate. To address these issues, we introduce a \textit{a contrario} detector \cite{ammar_-contrario_2013}\footnote{We let the reader refer to \cite{ammar_-contrario_2013} for a detailed description of the {\em a contrario} setting.}. It works at a 4-hour time scale and addresses the early detection of abnormal vessel behaviors. We divide the map into small cells $C_i$. In each cell, we calculate the mean $m_i$ and the standard deviation $std_i$ of the $\log p(x_t|x_{1:t-1},z_{1:t-1})|_{x_t \in C_i}$ using the tracks in the validation set. Any evolution $p(x_t|x_{1:t-1},z_{1:t-1})$ at timestep $t$ of an AIS track will be considered as an abnormal evolution if its log-likelihood is much lower than the distribution of other log-likelihoods in the same cell. The \textit{a contrario} detection detects if an arbitrary segment is abnormal based on the number of abnormal evolutions in this segment and its length.

\subsection{Vessel type identification submodel}
\label{secVesselTypeIdentification}

The third task addressed by our model is the identification of the vessel type from its AIS-derived trajectory data. It may be noted that the vessel type should be one of the attributes included in AIS messages. However, not all vessels send their static messages. Some may even send on purpose a false vessel type in AIS messages. A Vessel type identification submodel is then an important tool to detect suspicious behaviors.

Different types of vessels usually perform specific behaviors, which may differ among others in terms of geographical zones, speed patterns, etc. For example, tankers normally follow maritime routes (usually straight lines between two maritime waypoints \cite{pallotta_vessel_2013}), their average speed is relatively low, about 12-15 knots, whereas passenger ships have relatively high average speed, about 20-25 knots. If a vessel declares itself as type ``A'' but performs a maneuver of type ``B'', it is likely that it may carry out illegal activities. 

In this study, we design a Vessel type identification submodel at a 1-day time scale. 
This submodel explores a Convolutional Neural Network (CNN).
The input of this CNN is a $H$x$D$ matrix, whose columns are the hidden regimes (dimension $H$), and $D$ is the number of timesteps in one day (144 in this case). Because the hidden regime is regularly time-sampled, this configuration applies directly. 



\section{Experiments and Results}
\label{secExperimentsResults}

We implemented the proposed framework for a three-task model, addressing respectively vessel trajectory reconstruction, abnormal behavior detection and vessel type identification, in the Gulf of Mexico and the abnormal behavior detection off Brittany coast in the Ushant zone\footnote{The Tensorflow code and the datasets are available at https://github.com/dnguyengithub/MultitaskAIS}. The Ushant water is the entrance to English channel, this region is interesting to maritime surveillance because of its separation scheme and the heavy traffic there. The Gulf of Mexico is relatively large compared to the case-study regions considered in previous studies \cite{pallotta_vessel_2013}, \cite{laxhammar_anomaly_2008}, \cite{kowalska_maritime_2012}. This region involves multiple vessel types and activities of vessels. It comprises big ports, fishing zones, oil platforms and dense maritime routes. Overall we considered AIS data from January to March 2017 off Brittany coast in the  Ushant zone (2,021,236 AIS messages) and from January to March 2014 in the Gulf of Mexico (180,344,817 AIS messages). 

\subsection{Preprocessing}

For the pre-processing step, first, infeasible speed or infeasible position messages were removed from the set. To handle the problem of very long sequence when working with RNNs, we split vessels' tracks into subtracks of from 4 to 24 hours. 
From now on, in this paper, vessel tracks refer to such subtracks. 
We also removed tracks whose speed is smaller than 0.1 knots for more than 80\% of the time (at anchor or moored vessels).

One objective of the proposed architecture is to deal with irregularly-sampled data. However, we need regularly sampled data to train the model first. In a layman's term, the Embedding block must see how regularly sampled AIS tracks should be and learn their characteristics, after that (after being trained), it with generate regularly-sampled data from irregularly-sampled ones. Therefore, for the training set, we only chose tracks whose the maximum time interval between two successive received AIS messages is 1 hour, then used constant velocity model to create regularly time-sampled AIS tracks at 10-minute time scale. By doing this, the intervals between two successive AIS messages are small enough that the errors in the estimation of the constant velocity model do not effect our model too much. 

\subsection{Embedding block calibration}

We implemented the Embedding block by a VRNN whose the RNN is a single-layer LSTM, the distributions $p(x_t|z_t, h_t)$, $p(z_t|h_t)$, $q(z_t|x_t, h_t)$ are fully connected networks with one hidden layer of the same size of the LSTM's. $p(x_t|z_t, h_t)$ is binomial, $p(z_t|h_t)$ and $q(z_t|x_t, h_t)$ are Gaussians. The network was trained with stochastic gradient descent using Adam optimizer \cite{kingma_adam:_2015}, learning rate of 0.0003.

There is a trade-off between the resolutions of AIS features and the size of the network when choosing the length of the ``four-hot encoding''. If the resolutions are too high, the ``four-hot'' vector will be too long, requires a big hardware memory and computational power; if the resolutions are too low, we lose information. We set here the resolution of the latitude and longitude coordinate at about 1 km, the resolution of SOG at 1 knot and the resolution of COG at 5$\degree$. These resolutions are fine enough for almost all the maritime safety, security and efficiency tasks. For example, with this setting, the uncertainty zone of vessel's position is about 1kmx1km, small enough for position-related tasks. 

The choice of the dimension of hidden regime effects the modeling capacity of the Embedding block. As shown in Table \ref{tabLatentSize}, if the latent size is too small, the model can not capture all the variations of AIS data. In contrast, if the latent size is too big, the model becomes too bulky and overfitting. For the rest of this paper, we set the latent size at 400 for tests on the Gulf of Mexico dataset and at 100 for tests on the Brittany dataset (Ushant water).

\begin{table}[!t]
  \renewcommand{\arraystretch}{1.3}
  \caption{Log likelihoods of the Embedding block with different dimension settings (Gulf of Mexico dataset).}\label{tabLatentSize}
  \centering
  \begin{tabular}{| c | c | c | c |}
    \hline
	 Hidden regime 	& Number of 	&  Log likelihood 		& Log likelihood \\	 
     dimension 		& parameters	&  on training set		& on  test set  \\
    \hline
	200			&  1 605 402	& -7.592710 			& -7.678684		\\
   \textbf{ 400 }		&  \textbf{5 129 202}	& \textbf{-6.557936}  			& \textbf{-7.520255} 	\\
    500    		&  7 611 102    & -6.130078				& -7.690255     \\
    \hline
  \end{tabular}
\end{table}



\subsection{Vessel trajectory construction}

\begin{figure*}[!t]
  \centering
  \includegraphics[width=179mm]{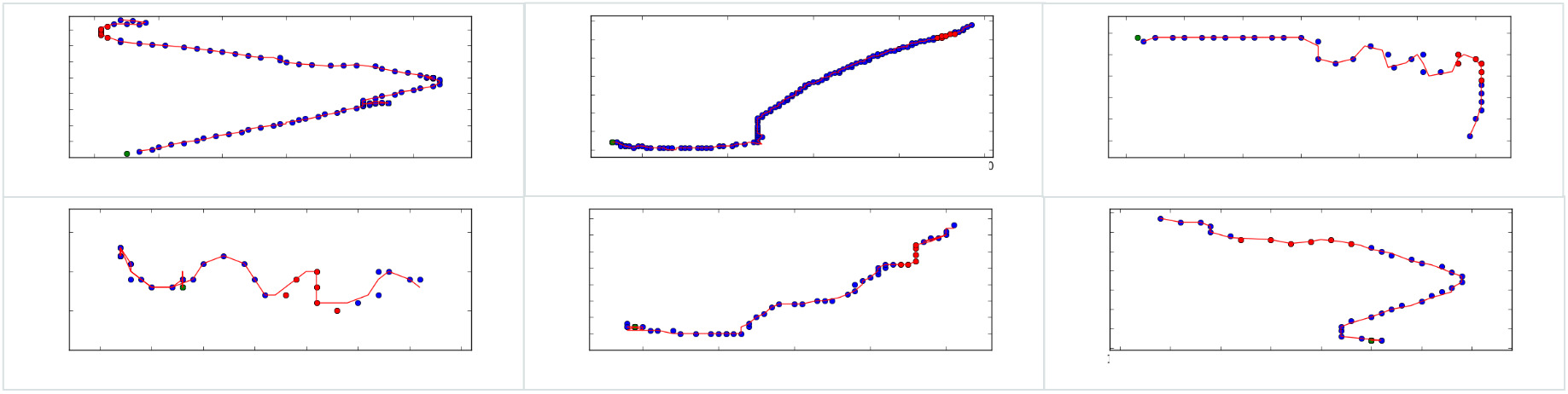}
  \caption{Trajectory reconstruction examples using the proposed model. Blue dots: received AIS messages; red dots: missing AIS messages; red lines: trajectories reconstructed by our model.} \label{figGoodRecons}
\end{figure*}

We deleted a 2-hour segment from each AIS track then used the Vessel trajectory construction layer to reconstruct this segment. The maritime contextual information learned by the Embedding block gave the model the ability to reconstruct some complex trajectories like those on the top right and bottom left of Fig. \ref{figGoodRecons}. These constructions can not be achieved by interpolation methods such as linear or spline interpolation. 

The performance of this layer depends strongly on the maritime contextual information extracted by the Embedding layer. If the extraction is good, the model can predict complicated patterns like those shown in Fig.~\ref{figGoodRecons}. However, in zones whose the vessel density is low, or in zones where the behaviors of vessels are too complicated for the Embedding layer to learn, the construction layer completely fails to estimate the positions of vessels. In these cases, we use constant velocity method. The switch between particle method and constant velocity method is automatic, because the model knows when the Embedding layer can not extract the maritime contextual information (based on the value of the probability $p_t(x_t|x_{1:t-1},z_{1:t-1})$).

\subsection{Abnormal behavior detection}

We divided each dataset into 3 sets: a training set to train the model, a validation set to calculate the the $mean$ and $std$ of the log probability, and a test set to test the anomaly detection. The proportion of the 3 sets was 60/30/10. Although the training sets were used for learning the normalcy model, we did not do data cleaning, {\em i.e.} the training sets themselves may contain abnormal trajectories. Our framework relies on probabilistic models and implicitly assumes that abnormal trajectories are rare events, that is to say that the probability mass at these trajectories would be very low.

\begin{figure*}[!t]
  \centering
  \includegraphics[width=\textwidth]{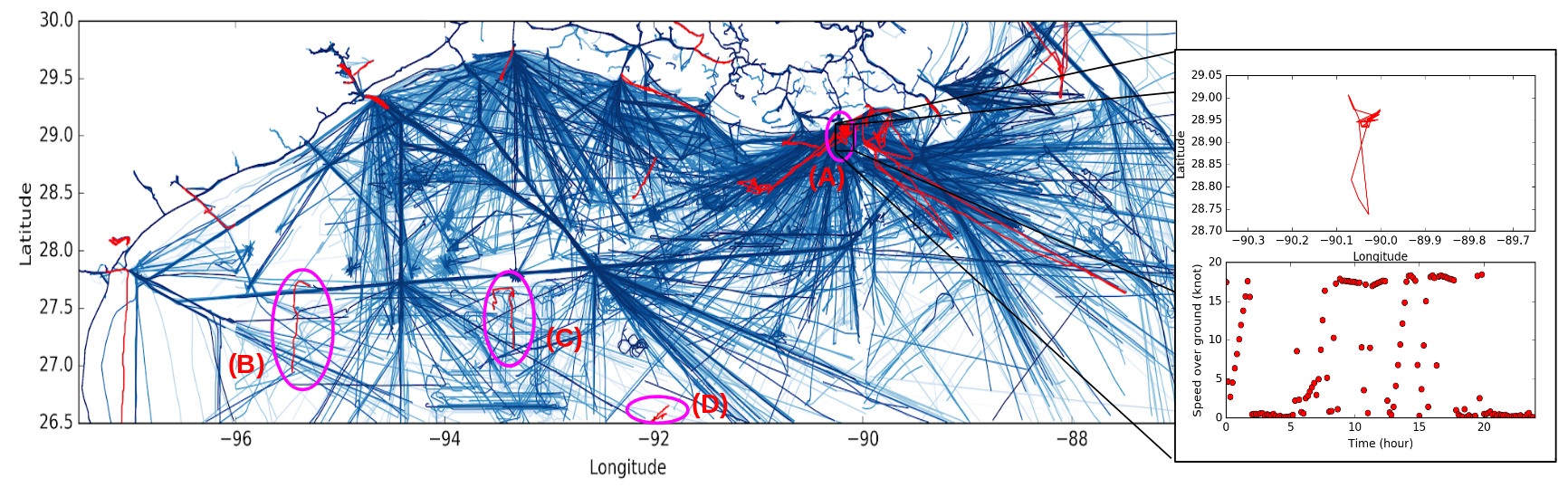}
  \caption{Detection of abnormal behaviors using global thresholding (Gulf of Mexico dataset). Blue: tracks in the training set (which itself may contain abnormal tracks); red: abnormal tracks detected in the test set. We highlight four examples: a track with an abnormal speed pattern \textbf{(A)} ii), two tracks with abnormal trajectory shapes from others' in the same region \textbf{(B,C)} iii) a track in a low-density area (abnormal zone) \textbf{(D)}.} \label{figAnomalyGlobal}
\end{figure*}

We report the outcome of the anomaly detection submodel when using global threshold detection on the Gulf of Mexico dataset in Fig.~\ref{figAnomalyGlobal}. 
A track will be detected as abnormal if its shape is unusual, its speed pattern is rare, or it appears in an abnormal region, etc. Each type of these anomalies corresponds to a signature of trajectory data, like geographical pattern, geometric pattern, speed and course distribution, etc. These signatures will be presented in Section \ref{secExplanations}. 
 

\begin{figure*}[!t]
  \centering
  	\subfloat[Mean of the log-likelihood in each cell]
    {\includegraphics[width=70mm]{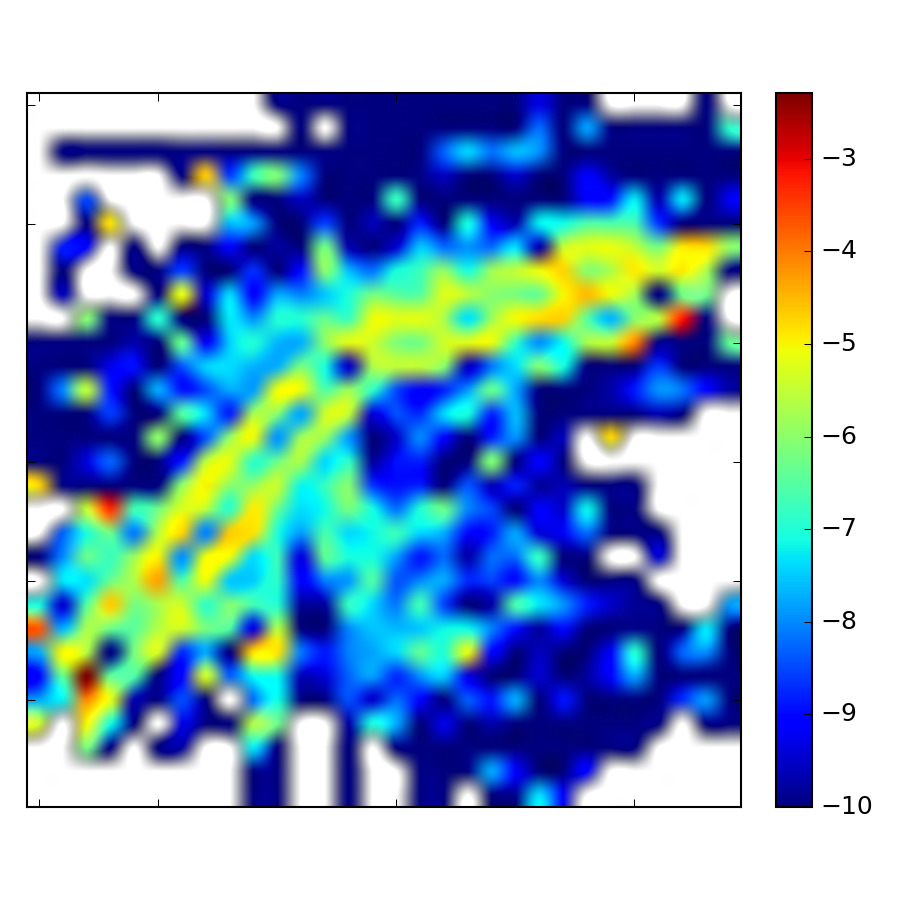}
 	\label{figMeanMap}}%
  \hfil
  	\subfloat[Std of log-likelihood in each cell]
    {\includegraphics[width=69.5mm]{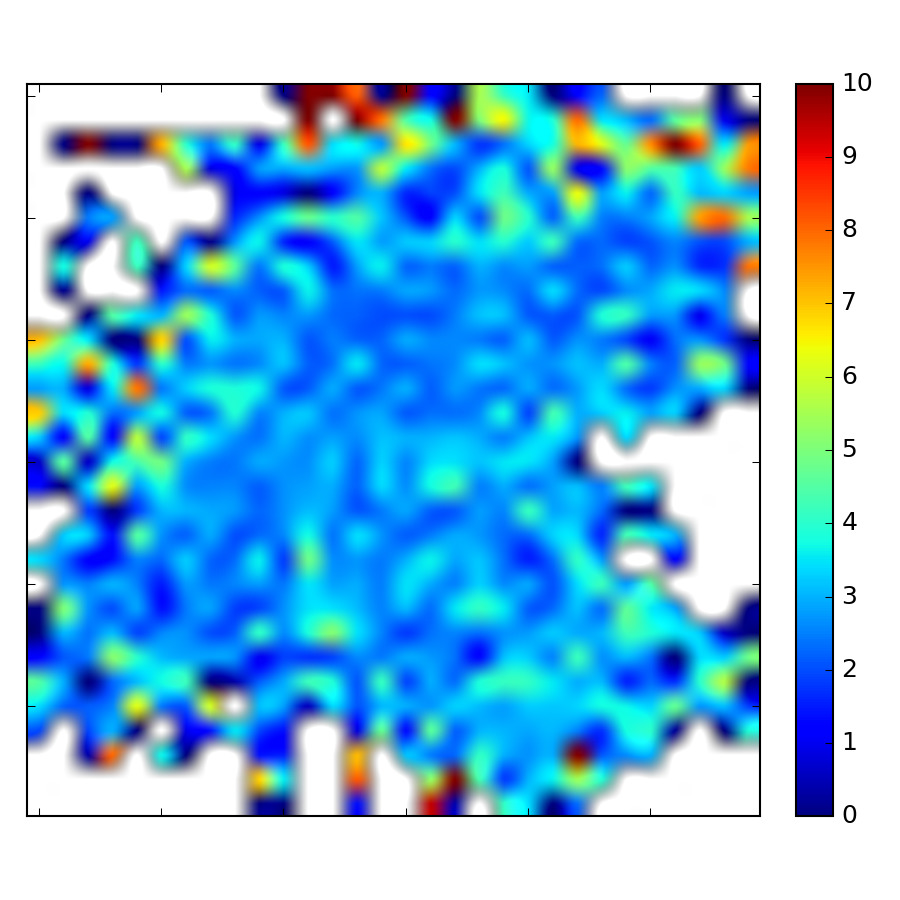}
 	\label{figStdMap}}%
  \centering
  \caption{Maps of the mean and the std of the log-likelihood of the trained model in each cell (Brittany dataset).}
  \label{figMeanStdMap}
\end{figure*}

\begin{figure*}[!t]
  \centering
  	\subfloat[]
    {\includegraphics[width=54mm]{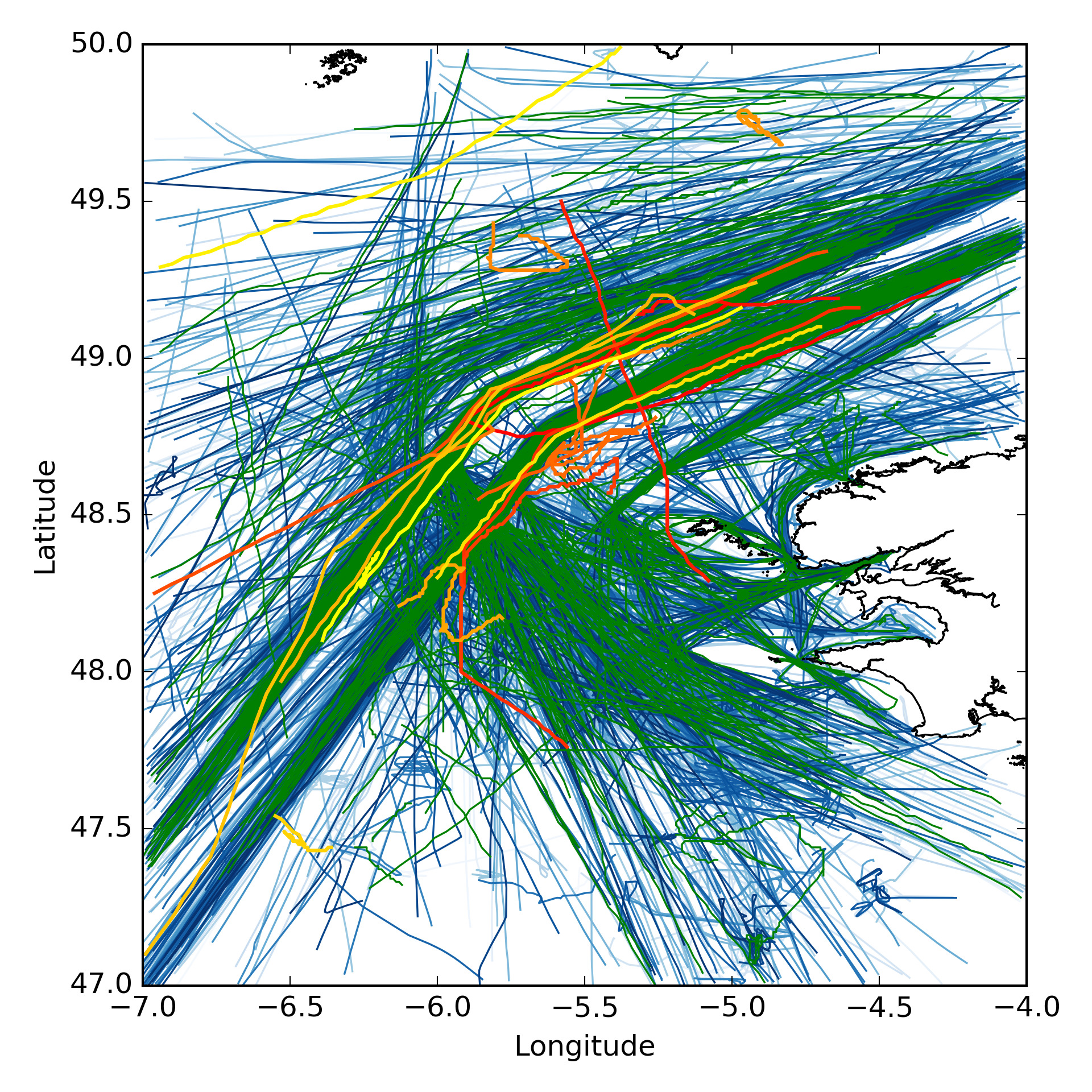}
 	\label{figContrarioA}}%
  \hfil
  	\subfloat[]
    {\includegraphics[width=54mm]{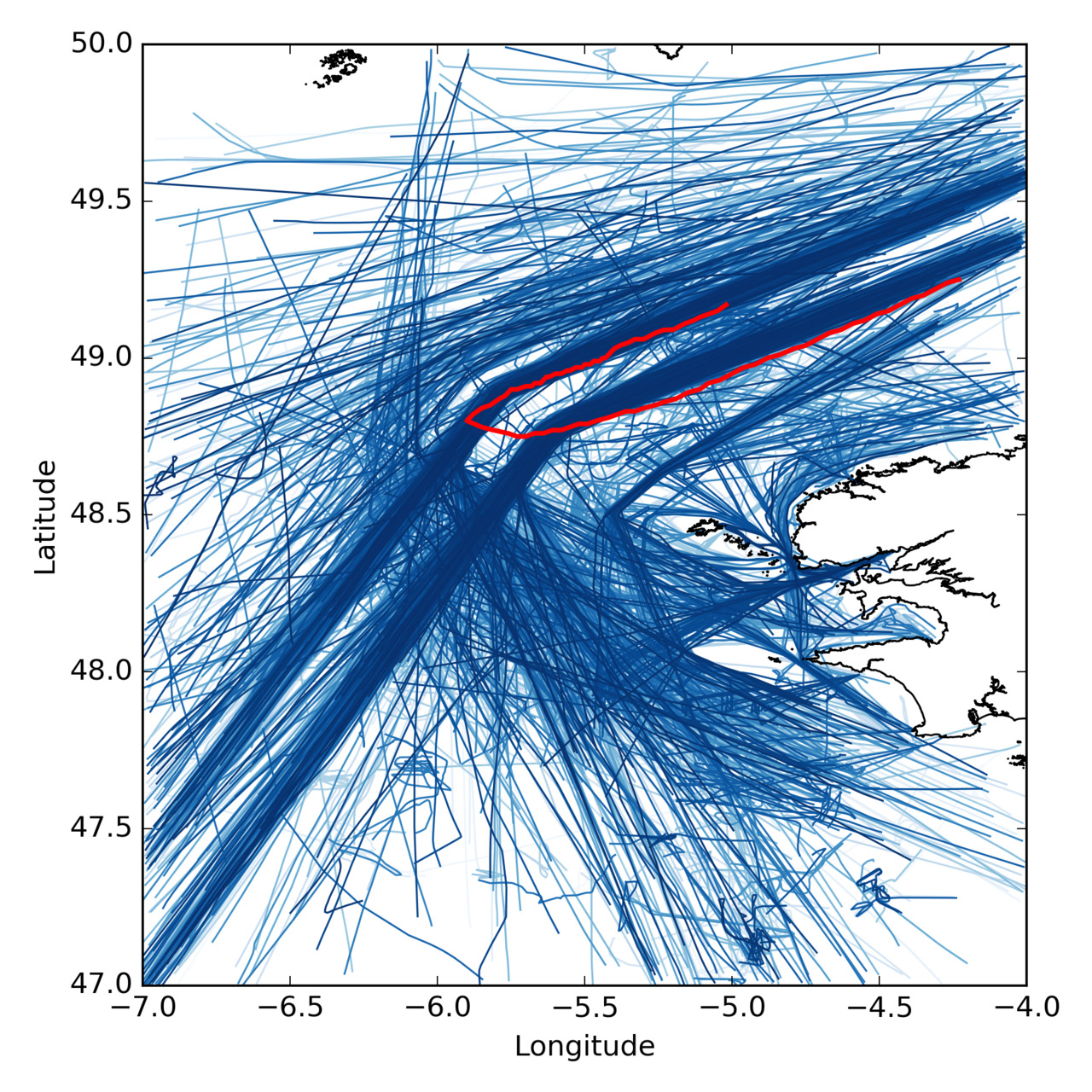}
 	\label{figContrarioB}}%
  \hfil
  	\subfloat[]
    {\includegraphics[width=54mm]{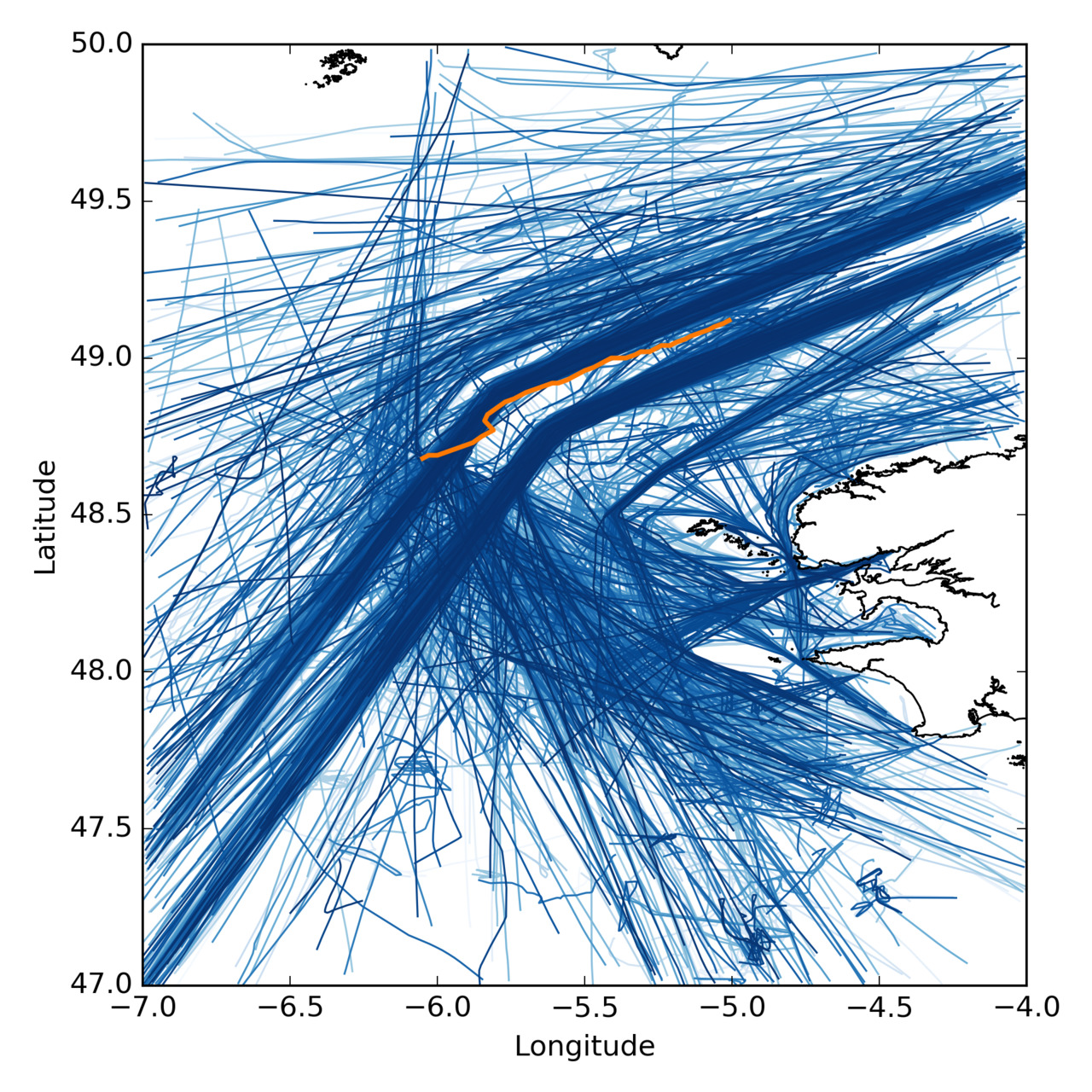}
 	\label{figContrarioC}}%
  \hfil
  	\subfloat[]
    {\includegraphics[width=54mm]{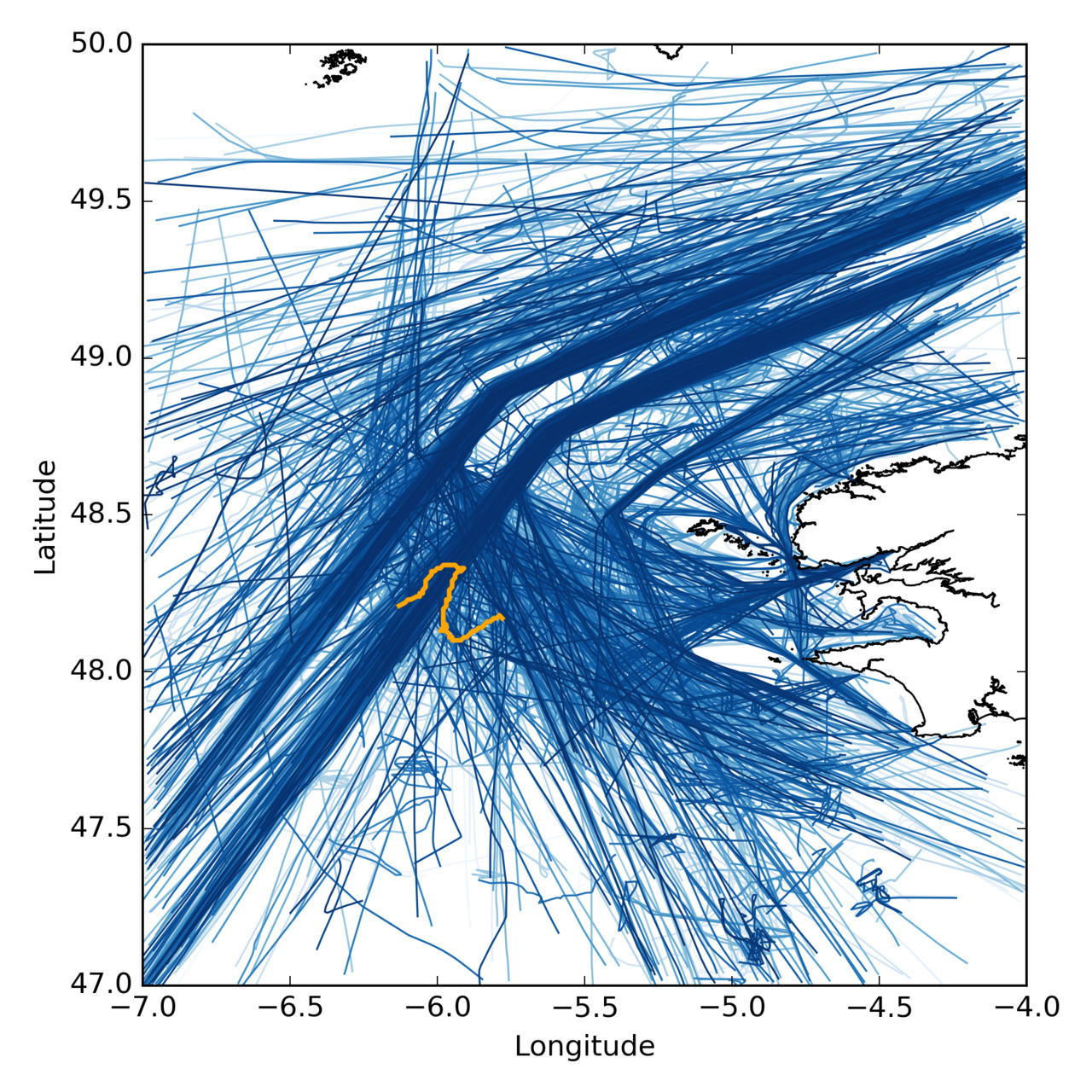}
 	\label{figContrarioD}}%
  \hfil
  	\subfloat[]
    {\includegraphics[width=54mm]{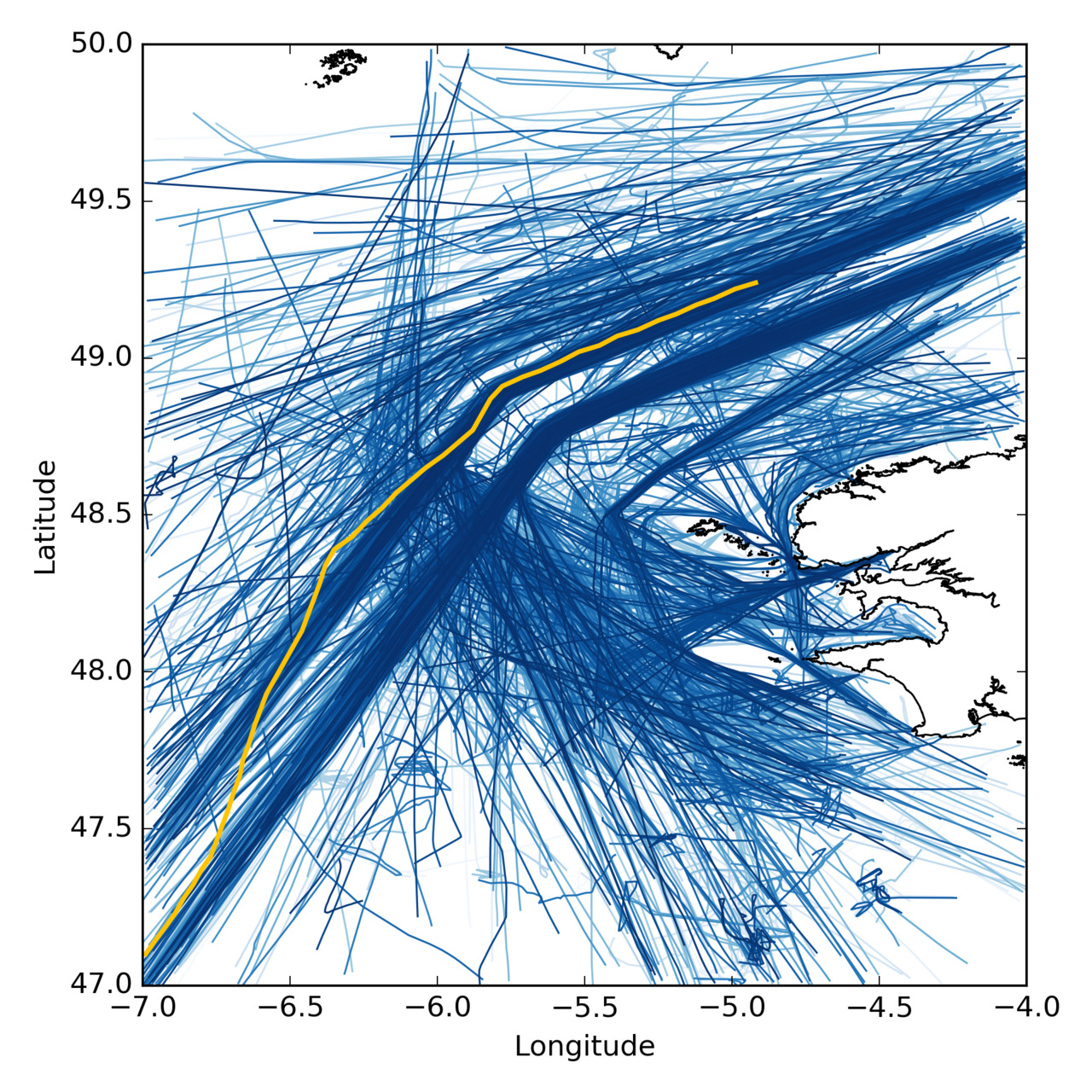}
 	\label{figContrarioE}}%
  \hfil
  	\subfloat[]
    {\includegraphics[width=54mm]{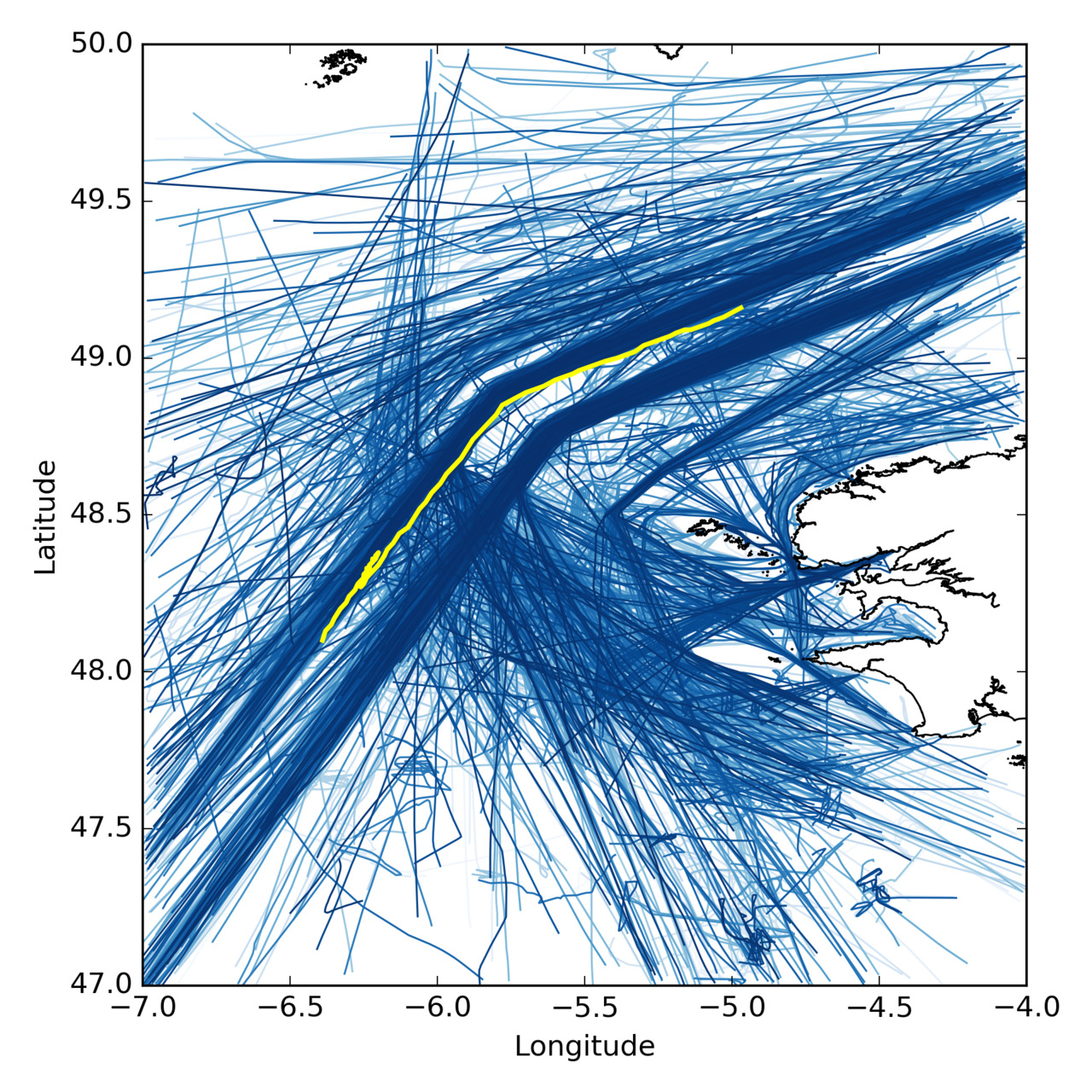}
 	\label{figContrarioF}}%
  \centering
  \caption{Abnormal tracks detected by the proposed \textit{a contrario} model (Brittany dataset). (a) All tracks detected in the test set; blue: tracks in the training set; green: normal tracks in the test set; other colors: abnormal tracks in the test set. (b) Abnormal U-turn. (c-d) Divergences from maritime route. (e) Abnormal route change. (f) Abnormal double-U-turn.}
  \label{figAnomalyContrario}
\end{figure*}

For the \textit{a contrario} detection, we split the ROI into small cells of 10kmx10km. The maps of the mean and the standard deviation of the log-likelihood on the Ushant dataset are shown in Fig. \ref{figMeanStdMap}. We can see that the log-likelihood strongly depends on geographical region, global thresholding would not work. 
On the mean map, there are some lines/curves of high value, they are the maritime routes. On the maritime routes, the vessel density is high, vessels performs simple and similar maneuvers, so the model can learn these patterns easily. On the other hand, in regions where the vessel density is low, or the behaviors of vessels are too complicated, the identification of abnormal behaviors appears more complex and may require larger training datasets. Detection examples and their corresponding interpretation are shown in Fig. \ref{figAnomalyContrario}.

\begin{figure*}[!t]
  \centering
  \includegraphics[width=\textwidth]{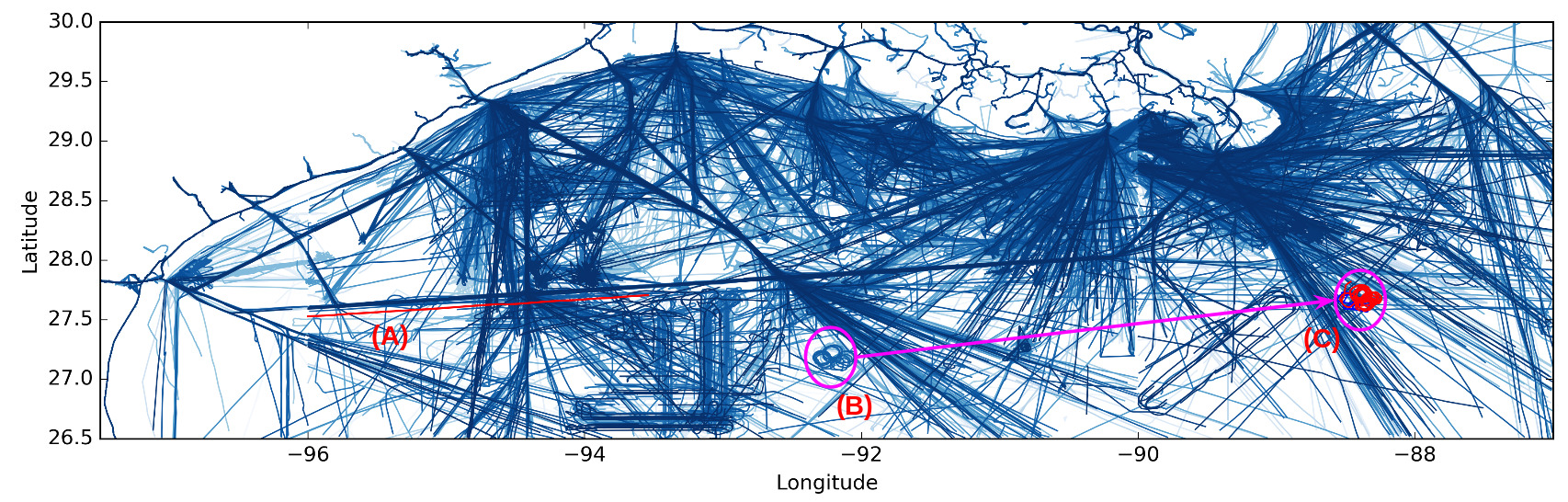}
  \caption{Example of the \textit{a contrario} anomaly detection on simulated dataset (Gulf of Mexico dataset). The circle-shaped tracks in zones \textbf{(C)} were simulated by translating from \textbf{(B)}; \textbf{(A)} is a detection of a divergence from maritime route.} 
  \label{figAnomalySimulation}
\end{figure*}

To evaluate the consistency of the \textit{a contrario} model, we tested this detector for simulated abnormal examples. We translated a normal track out of maritime routes to simulate the divergence from a given route (zone (A)) and translated circle-shaped tracks in zone (B) to zone (C) in Fig. \ref{figAnomalySimulation} to verify that some specific patterns of vessels' maneuvers should appear in their specific zones. Experiment shows that the model can detect the divergences if the distance to the maritime route is far enough (10km) and it detects 9 over 13 circle-shaped tracks in zone (C).

In comparison to methods in the literature, our method has several benefits:
\begin{itemize}
	\item It can detect abnormal patterns that are detected in state-of-the-art methods, such as the double-U-turn detection reported in Fig. \ref{figContrarioF} and also illustrated in \cite{pallotta_vessel_2013}.
    \item Methods like those in \cite{pallotta_vessel_2013} and \cite{mascaro_anomaly_2014} first assign a track to a maritime route, then compare the similarity between this track with the those in the corresponding route to decide whether this track is normal. However, it is very difficult to link tracks like the one in Fig. \ref{figContrarioD} to a maritime route. Therefore, our model which does not require the prior identification of maritime routes appears more generic and robust.   
    \item Our model relaxes strong assumptions. In \cite{pallotta_vessel_2013}, the authors assumed that the probability of observing a feature vector ($[lon_t,lat_t,SOG_t, COG_t]^T$ in their case) of a vessel at the time $t$, given its position and assigned route was independent. This assumption neglects the fact that AIS streams provide sequential data, feature vectors of a vessel's track are related to this vessel and interdependent. For instance, for such approaches the two branches of the ``U'' in Fig. \ref{figContrarioB} are normal. 
    \item Methods in the literature do not deal with irregularly time-sampling problem. For example, model in \cite{pallotta_vessel_2013} used sliding window to avoid incomplete tracks, and processed only the most recent points of the partially observed tracks. The vessel in Fig. \ref{figContrarioB} can outsmart this model by switching off its AIS transponder when performing the U-turn (which lasts about 30 minutes).
    \item In a complicated region like the Gulf of Mexico, all the methods based on DBSCAN \cite{mazzarella_knowledge-based_2015}, \cite{pallotta_vessel_2013} cannot apply since DBSCAN fails to extract effective waypoints. As shown in Fig. \ref{figAnomalySimulation}, in this area, vessels do not enter or exit the ROI at some specific zones, in consequence, DBSCAN can not detect entry and exist waypoints; beside that, a lot of vessels stop at sea for purposes (fishing for example), leads to false stationary waypoint detection by DBSCAN. 
\end{itemize}



\subsection{Vessel type identification}


We tested the Vessel type identification submodel with a set of 1800 AIS tracks of 4 types of vessels: cargo, passenger, tanker and tug.     

We compared the performance of our model with the one of other types of neural networks: CNNs and LSTMs (which are currently the state-of-the-art for time series classification). To simulate the missing data phenomenon in AIS streams, we deleted a 2-hours segment in each AIS track. Constant velocity model was used to fill the missing points for CNN model. We tested the LSTM networks with and without the ``four-hot encoding'' layer to show the benefit of this presentation. For each type of architecture we tried several configurations and report the best result. 

The results are shown in Table.~\ref{tabClassificationResult}. First, the poor performance of LSTMs without the ``four-hot encoding'' layer shows the relevance of this presentation for disentangling the explanatory information in continuous feature spaces of AIS messages' attributes.  Second, we can see that the proposed model achieved comparable performances with those of the state-of-the-art methods. It is because the embedding layer can provide a solid regular series of hidden regimes despite irregular time sampling in AIS streams. In addition to the slight improvement of the classification performance (from 87.43\% to 87.72\%), the proposed model also significantly reduces storage redundancies and computational requirements when doing each task separately, which is highly beneficial in an AIS big data context.

\begin{table}[!t]
  \renewcommand{\arraystretch}{1.3}
  \caption{Classification results.}\label{tabClassificationResult}
  \centering
  \begin{tabular}{| l | c | c | c |}
    \hline
    Model		& Precision 		& Recall 		& F1-score \\
    \hline
	LSTM			&  47.51\%			& 64.11\%  			& 52.08\% 	\\
    LSTM\_4-hot 	&  \textbf{88.04\%}	& 87.16\%  			& 87.43\% 	\\
    CNN    			&  83.83\%          & 84.06\% 			& 83.75\%   \\
    \textit{\textbf{VRNN-CNN}}	&  \textit{88.00\%} & \textit{\textbf{87.67\%}}  & \textit{\textbf{87.72\%}}	\\
    \hline
  \end{tabular}
\end{table}


\section{Insights on the considered approach}
\label{secExplanations}

In this section, we further discuss the key features of the considered approach with respect to state-of-the-art approaches. Overall, AIS vessel tracks (and trajectory data in general) may be characterized according to the following features:
\begin{itemize}
    \item Time evolutions in terms of vessel position, speed and course;
	\item Geographical patterns (where is the vessel?);
    \item Geometric patterns (what is the shape of the track?);
    \item Speed and course distributions;
    \item Spatial-temporal patterns, called \textit{``phase patterns''} (moves fast in specific zones and slowly in other zones for example). 
\end{itemize}
We discuss below how our approach addresses the learning of these key features.


In time series, different features change at different temporal and spatial scales \cite{bengio_representation_2013}. The proposed model learns these features from different points of view at different scales. At micro-scales, it learns the evolutions of the trajectories, {\em e.g.} with this historical information, in 10 minutes, vessel ``V'' seems to appear in zone ``Z'', maintain its speed around ``S'' knots. These evolutions are modeled by the distribution $p(x_t|x_{1:t-1},z_{1:t-1})$. At macro-scales, the model tends to learn the patterns of the entire AIS tracks. 

\begin{figure}[t!]
  \centering
  \includegraphics[width=82mm]{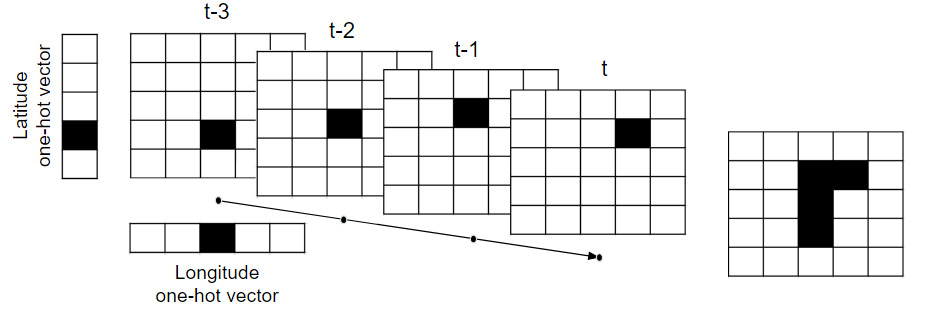}%
  \caption{Geometric patterns appear by summing up one-hot vectors of latitude and longitude coordinates.} \label{figOnehotAccum}
\end{figure}

Viewing the ``four-hot'' representation as an image-based representation of a track seems relevant to understand how our model can learn complex space-time patterns. More precisely, the one-hot vectors of the latitude/longitude coordinates of AIS messages indicate the rows/columns of the pixels in the image, respectively. Hence, if we cumulate these two one-hot vectors over a given time period, we build an image-based representation, which describes the geometric pattern of the vessel track (Fig. \ref{figOnehotAccum}).

\begin{figure}[t!]
  \centering
  \includegraphics[width=78mm]{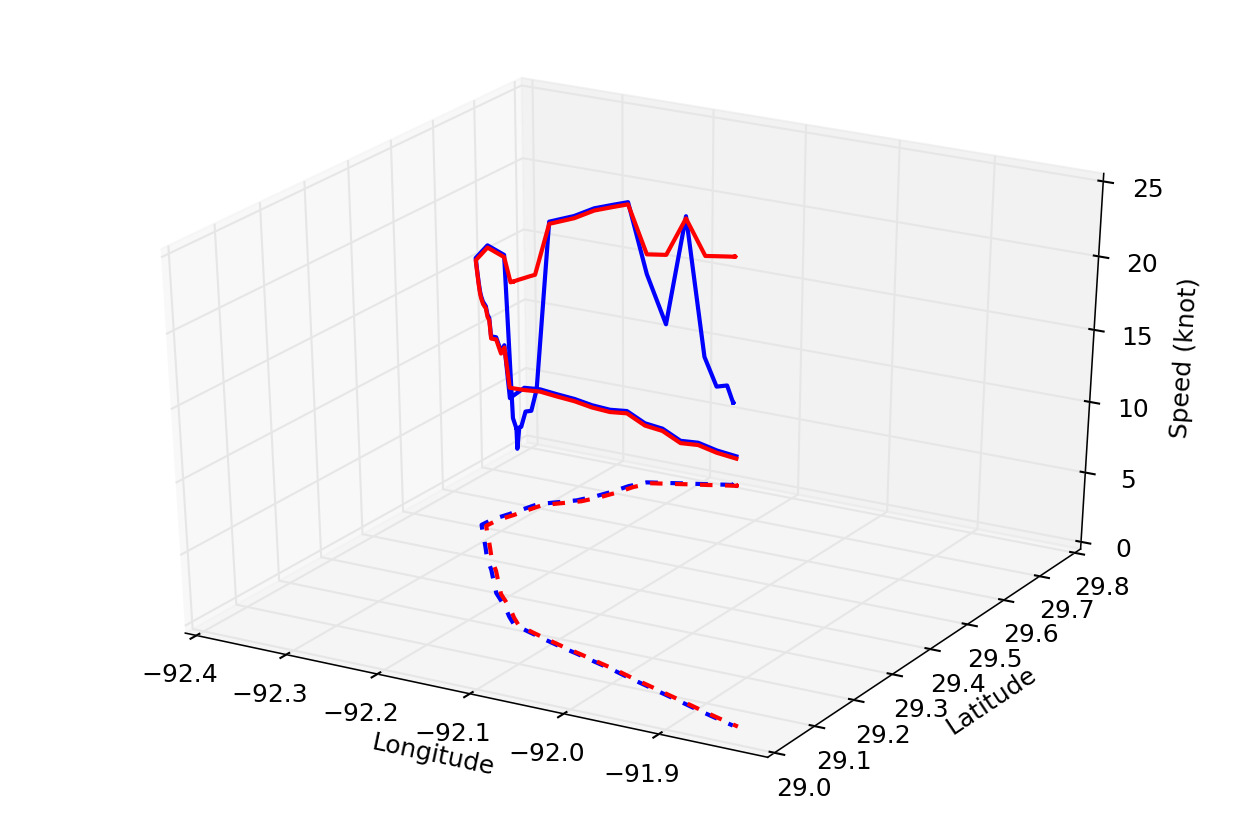}
  \caption{Illustration of phase patterns. We report the two examples of two AIS tracks of the processed dataset (red and blue). The solid lines are 3D curves (latitude, longitude and speed time series) reflect the phase pattern, whereas the dash curves (latitude and longitude time series only) reflect the associated 2D geometric patterns, which can not reveal the observed phase patterns.
  }
  \label{figPhasePattern}
\end{figure}

The proposed model is not translation-invariant
and can learn spatial patterns and the geographical distributions of vessel tracks, {\em i.e.} a given type of tracks should appear in zone ``A'' and not in an other zone ``B''. The phase patterns, on the other hand, reflect the correlations between temporal features and spatial ones. One typical example of phase pattern in trajectory data is the speed-position correlation, {\em e.g.} the average speed of vehicles on highway is higher than the one in urban area.  Methods that use only the positions (longitude and latitude coordinates) to model trajectory and consider the speed as the first-order derivative of the positions can not capture this information. For example, the two tracks depicted in Fig. \ref{figPhasePattern} are two examples from the processed dataset. They are similar in terms of spatial patterns, but different in terms of phase patterns, space-speed time series. Despite inter-individual variabilities, these two tracks exhibit in some regions low vessel speed and high vessel speed in other regions. 


These different aspects are similar to the wave-particle duality in physics, where the patterns correspond to the wave properties and the evolutions correspond to the particle properties.






\section{Conclusions and perspectives}
\label{secConclusionsPerspectives}


In this paper, we proposed a novel deep-learning-based scheme for maritime traffic surveillance from AIS data streams. Stated within a probabilistic framework using Variational RNN, our approach overcomes strong limitations of state-of-the-art methods to jointly address multi-task issues, namely abnormal behavior detection, trajectory reconstruction and vessel type identification, on a regional scale, that is to say for datasets of spatially-heterogeneous datasets of tens or hundreds of millions of AIS data. More precisely, we tackled three main drawbacks of state-of-the-art approaches: 
\begin{itemize}
	\item First, we relax strong assumptions usually considered such as a finite number of behavioral categories (or hidden regimes) \cite{holst_stattistical_2016}, \cite{gaspar_analysis_2016}. 
	\item Second, by using VRNN, we can capture the maritime contextual information while avoiding problems that may be encountered if doing clustering.
    \item Third, the Embedding block in our model can deal with noise and irregularly time-sampling of AIS data streams. Besides, the Embedding block also results in an efficient compression of the behavioral information conveyed in data, which avoids making accessible the entire training dataset for the operational use of the trained model. This appears critical for an operational big-data-compliant AIS system.
\end{itemize}
We also discussed the key aspects of the considered trajectory data representation, which is embedded in the considered VRNN framework. 


Beyond benchmarking issues for large-scale datasets,  including the evaluation of the ability of the proposed approach to scale up to global AIS data streams, the fusion with other sources of information available in the maritime domain could be a promising solution. Weather and ocean conditions, such as sea surface winds and currents, are two important factors that effect the behaviors of vessels. The exploitation of such variables should further constrain the considered VRNN framework and improve its representativity. The inference of behavioral models in low-density areas might require specific investigations in future studies, for instance some type of regularization.



\section{Acknowledgements}
\label{secAcknowledgements}

This work was supported by public funds (Minist\`ere de l'Education Nationale, de l'Enseignement Sup\'erieur et de la Recherche, FEDER, R\'egion Bretagne, Conseil G\'en\'eral du Finist\`ere, Brest M\'etropole) and by Institut Mines T\'el\'ecom, received in the framework of the VIGISAT program managed by ``Groupement Bretagne T\'el\'ed\'etection'' (BreTel).

The authors acknowledge the support of DGA (Direction G\'en\'erale de l'Armement) and ANR (French Agence Nationale de la Recherche) under reference ANR-16-ASTR-0026 (SESAME initiative), the labex Cominlabs, the Brittany Council and the GIS BRETEL (CPER/FEDER framework).

We also would like to thank MarineCadastre for the Gulf of Mexico AIS dataset and Collecte Localisation Satellites as well as Erwan Guegueniat for the Brittany AIS dataset.

\bibliographystyle{IEEEtran}
\bibliography{Zotero}

\end{document}